# Optimal Function Approximation with Relu Neural Networks

Bo Liu　　　　　　　　　　Yi Liang

Faculty of Information Technology, Beijing University of Technology, Beijing, China

liubo@bjut.edu.cn;　　　　liangyi@emails.bjut.edu.cn

**Abstract.** We consider in this paper the optimal approximations of convex univariate functions with feed-forward Relu neural networks. We are interested in the following question: what is the minimal approximation error given the number of approximating linear pieces? We establish the necessary and sufficient conditions and uniqueness of optimal approximations, and give lower and upper bounds of the optimal approximation errors. Relu neural network architectures are then presented to generate these optimal approximations. Finally, we propose an algorithm to find the optimal approximations, as well as prove its convergence and validate it with experimental results.

## 1 Introduction

Deep neural networks have achieved state-of-the-art performance in various fields of artificial intelligence such as computer vision, speech recognition and natural language processing. Despite these empirical successes, the theoretical understanding of deep learning remains elusive. One of the most important theoretical aspects about deep neural networks is the expressive power, which describes their ability to approximate functions.

It is well-known since late 80s that large networks with a single hidden layer can already approximate any continuous function on a compact domain to arbitrary accuracy [1, 2]. This result is referred to as the Universal Approximation Theorem. Due to the revival of deep learning in recent years, there have been some works on function approximation with deep neural networks, and various approximation constructions have been proposed, showing their universal approximation capabilities [3~7] and advantage of deep neural networks over shallower ones in approximation efficiency [8~19].

We concern the optimal approximation problem in this paper: given a continuous function and the maximal number of linear pieces output by a Relu neural network, what is the minimal possible approximation error? This optimal approximation problem is of fundamental theoretical importance and tells us how well at most one can approximate, and may have practical implications for some other deep learning problems as well, such as the limit of network compression without sacrificing accuracy. Although many approximation constructions have been proposed so far, such as using trapezoid [4] or spike [5] shaped units to approximate functions in local regions, they are not optimal in the sense that in all possible constructions given a fixed number of approximating linear pieces, their approximation errors are not minimal. In contrast, we seek for approximations in this work that are optimal and independent of any specific approximation constructions.

We study the optimal approximations of univariate convex functions with Relu neural networks, and have discovered the optimal approximations uniquely determined by the intrinsic nature of target functions. More specifically, we have made the following contributions in this paper:

1. We give and prove the necessary and sufficient conditions for optimal approximations of convex univariate functions with piecewise linear functions. The uniqueness of optimal approximations is proved as well.
2. Relu neural network architectures are designed to generate the optimal approximations.
3. Lower and upper bounds of optimal approximation errors are presented, both in terms of number of approximating linear pieces and network size.
4. An optimization algorithm is proposed to find optimal approximations. Its convergence is proved mathematically and validated with experimental results.

The paper is organized as follows. Section 1.1 is related work. In section 2, we introduce some background knowledge and notations. In section 3, we give and prove the optimality conditions, and derive the lower and upper bounds of the optimal approximation errors. Section 4 presents our Relu neural network architectures that output the desired optimal approximations. Section 5 introduces an algorithm to find optimal approximations, along with the convergence proof and experimental results to demonstrate its effectiveness. Finally, we give our conclusion and point out promising future directions.

**1.1 Related work**

Some local approximation constructions are proposed to represent general functions, including piecewise linear trapezoid [4, 7], piecewise linear spike-shaped unit [5], max-min string of affine pieces [3]. Comparing with them, our construction achieves optimal approximation.

There are some other constructions that first approximate polynomials [8, 9, 10, 11] and then use them as media to approximate more general functions. It has also been shown that comparing with these deep Relu network constructions, shallow networks have to be exponentially wider in order to achieve identical approximation accuracy. This comes from the fact that for sufficiently smooth functions there exist lower bounds of approximation errors that are determined by the number approximating linear pieces, which in turn are dominated by depth. In this work, besides lower and upper bounds, more importantly, we also give the exact expression of optimal approximation errors.

To show the important role of depth in enabling neural networks' expressive power, it has also been proved that some hard functions, such as sawtooth discrimination function for binary classification [13, 14], indicator functions of balls and ellipses [10, 16], $L_1$ radial function [10], and smoothly parametrized family of zonotope functions [17], can be represented easily by deep Relu networks, and in contrast cannot be approximated to certain accuracy by shallow Relu networks unless they are exponentially wider.

[18, 19] consider the expressive power of deep networks for functions with a compositional nature. Some authors have also considered the expressive power of deep networks other than standard feed-forward models, such as convolutional neural networks (CNNs) [6], ResNets [7, 20], and recurrent neural networks (RNNs) [21]. The expressive power of deep networks has also been approached from perspectives other than approximation errors, such as the number of linear regions [22, 23], trajectory length [24] and curvature [25] at higher layers.

## 2 Preliminaries

We use Relu neural networks to approximate strict convex univariate functions $f(x)$, that is, the second-order derivatives of $f(x)$ are positive: $f''(x) > 0$.

We consider feed-forward neural networks that are composed of layers of neurons, with each neuron computing a function of the form $x \mapsto \sigma(w^\top x + b)$, where $w$ is a weight vector, $b$ is a bias term and $\sigma$ is the Relu activation function defined as $\sigma(z) = \max(0, z)$. Let $\mathbf{W} = (w_1, w_2, \cdots, w_m)^\top$, $b = (b_1, b_2, \cdots, b_m)^\top$, and let $\sigma$ compute component-wise, we can define a layer of neurons as $x \mapsto \sigma(\mathbf{W}x + \mathbf{b})$. Denoting the output of the *i*th layer as $\mathbf{O}_i$, we can then define a Relu neural network of arbitrary depth recursively by $\mathbf{O}_{i+1} = \sigma(\mathbf{W}_{i+1}\mathbf{O}_i + \mathbf{b}_{i+1})$, where $\mathbf{W}_i$, $\mathbf{b}_i$ are the weight matrix and bias vector of the *i*th layer respectively.

The Relu activation function is piecewise linear with two pieces. When taking piecewise linear functions as input, the outputs of both affine transformation and Relu activation are still piecewise linear. Therefore, any neuron in the intermediate and output layers of a Relu neural network outputs a piecewise linear function. Approximating with Relu neural networks thus amounts to approximating with piecewise linear functions.

The $L_\infty$ approximation error is considered in this paper, defined as $sup_{x \in \chi}|f(x) - \tilde{f}(x)|$, where $\tilde{f}(x)$ is the function used to approximate target function $f(x)$, and $\chi$ is the domain of interest.

## 3 Optimal Approximation of Univariate Convex Functions
### 3.1 Necessary and sufficient conditions for optimal approximation

Given a univariate convex function *f(x)* defined in interval [*a*, *b*], and a piecewise linear function *f<sub>n</sub>(x)* with *n* linear segments to approximate *f(x)*, we are trying to answer the following question: what conditions do these linear segments must satisfy in order to achieve minimal approximation error? More formally, what segments will achieve the following optimal approximation error?

$$\Delta(f_n^*) \triangleq \inf_{f_n} \max_{a \leq x \leq b} |f(x) - f_n(x)|,$$

where $f_n^*(x)$ is the optimal piecewise linear function with *n* segments.

The segments partition interval [*a*, *b*] into *n* sub-intervals [*a<sub>i</sub>*, *b<sub>i</sub>*], *i*=1, 2…*n*, and in each [*a<sub>i</sub>*, *b<sub>i</sub>*], *f<sub>n</sub>(x)* is a single linear piece which we denote as $S_i$. We denote by $\Delta(S_i)$ the approximation error of a single segment $S_i$,

$$\Delta(S_i) \triangleq \max_{a_i \leq x \leq b_i} |f(x) - S_i(x)|.$$

The corresponding optimal segment is denoted as $S_i^*$, i.e., $\Delta(S_i^*) \triangleq \inf_{S_i} \Delta(S_i)$.

We now present our necessary and sufficient conditions for optimal approximation in the following theorem.

**Theorem 1**. *Given a strict convex univariate function f(x) and a Relu neural network generated piecewise linear function f<sub>n</sub>(x) with n linear segments to approximate f(x), the necessary and sufficient conditions for f<sub>n</sub>(x) to achieve optimal approximation are*

$$\Delta(S_1^*) = \Delta(S_2^*) = \cdots = \Delta(S_n^*) = \Delta(f_n^*) \tag{3.1}$$

*The optimal approximation is also unique.*

In order to prove Theorem1, we need two additional lemmas, i.e., Lemma 1 and Lemma 2.

**Lemma 1** (optimal approximation by a single segment). *For functions $f(x) \in C^2[a,b]$ with $f''(x) > 0$ ($a \leq x \leq b$), the optimal approximation error by a line segment S(x) is*

$$\Delta(S^*) = \frac{c-a}{2}[f'(c) - f'(d)], \tag{3.2}$$

*where c is determined by*

$$f'(c) = \frac{f(b)-f(a)}{b-a}, \tag{3.3}$$

*and d is determined similarly by*

$$f'(d) = \frac{f(c)-f(a)}{c-a}. \tag{3.4}$$

*Proof.* For optimal function approximation with polynomials, there is the Chebyshev theorem [26], which states that the optimal approximation of $f(x)$ by $1^{st}$-order polynomial $p(x)$ (i.e., a line segment) exists uniquely, and the error $p(x) - f(x)$ must assume optimal values with interleaving signs at at least 3 points

$$x_1 < x_2 < \ldots < x_N \quad (N \geq 3).$$

Assuming the optimal line segment is $S^*(x) = Ax + B$, according to Chebyshev theorem, there is at least one point in the interior region $(a, b)$ that assumes the optimal approximation error. Let c be such a point, thus it should be a stationary point of $S^*(x) - f(x)$,

$$S^{*\prime}(c) - f(c) = A - f'(c) = 0.$$

Therefore $A = f'(c)$. Since $f''(x) > 0$, $f'(x)$ increase monotonically, hence except for point $c$, $(S^*(x) - f(x))' = A - f'(x)$ cannot equal zero elsewhere in $(a, b)$. This indicates that there are no other points in $(a, b)$ achieving optimal error, and the remaining points that achieve optimal error must be the two endpoints. Putting together, there are totally three points $a$, $b$, $c$ achieving optimal approximation error with interleaving signs,

$$S^*(a) - f(a) = -[S^*(c) - f(c)] = S^*(b) - f(b). \tag{3.5}$$

Solving (3.5) gives

$$A = \frac{f(b)-f(a)}{b-a}, B = \frac{f(a)+f(c)}{2} - \frac{a+c}{2}A, \tag{3.6}$$

where $c$ is determined by

$$f'(c) = A = \frac{f(b) - f(a)}{b - a},$$

and $c$ is unique due to monotony of $f'(x)$.

We now compute the optimal approximation error $\Delta(S^*)$.

$$\Delta(S^*) = f(a) - S^*(a) = f(a) - (Aa + B) = f(a) - \left(Aa + \frac{f(a)+f(c)}{2} - \frac{a+c}{2}A\right) \tag{3.7}$$

By definition of point $d$ in (3.4),

$$f(c) = f'(d) \cdot (c - a) + f(a). \tag{3.8}$$

substituting (3.8) into (3.7), we get

$$\Delta(S^*) = f(a) - \left[f(a) + \frac{c-a}{2}f'(d) - \frac{c-a}{2}A\right] = \frac{c-a}{2}[f'(c) - f'(d)]$$

This completes the proof. □

**Lemma 2.** *For convex functions $f(x)$, if one enlarges [a,b] by moving the endpoint b right or moving a left, the optimal approximation error $\Delta(S^*)$ of $f(x)$ by a line segment $S(x)$ will increase. On the contrary, if one moves b left or a right, $\Delta(S^*)$ will decrease.*

*Proof.* We first prove the case of moving $b$. By (3.2), the differential of optimal approximation error caused by moving $b$ is

$$d(\Delta(S^*)) = \frac{dc}{2}[f'(c) - f'(d)] + \frac{c-a}{2}[df'(c) - df'(d)], \tag{3.9}$$

where $dc$ is the differential of point $c$ caused by movement of $b$. Using (3.4), we have

$$df'(d) = \frac{f'(c)dc \cdot (c-a) - [f(c)-f(a)] \cdot dc}{(c-a)^2}.$$

Substituting this expression into (3.9) yields

$$d(\Delta(S^*)) = \frac{(c-a)dc}{2}[f''(c) + \frac{f'(c)-f'(d)}{c-a} - \frac{f'(c)}{c-a} + \frac{f(c)-f(a)}{(c-a)^2}].$$

Using definition in (3.4) again, we get

$$d(\Delta(S^*)) = \frac{(c-a)}{2} \cdot f''(c)dc. \tag{3.10}$$

Now we establish the relationship between $dc$ and $db$. From definition in (3.3),

$$f''(c)dc = \frac{f'(b)db(b-a)-(f(b)-f(a))db}{(b-a)^2} = \frac{db}{b-a}[f'(b)-f'(c)]. \tag{3.11}$$

Combining (3.10) and (3.11) results in

$$d(\Delta(S^*)) = \frac{c-a}{2(b-a)} \cdot [f'(b)-f'(c)] \cdot db \tag{3.12}$$

Recall that $f''(x) > 0$ and $b > c$, thus $f'(b) - f'(c) > 0$, $d(\Delta(S^*))$ will have the same sign with $db$. Therefore the case concerning moving $b$ is proved.

For the case of moving endpoint $a$, we first flip the function $f(x)$ horizontally, which does not change the value of optimal approximate error. As a result, moving the left endpoint of flipped function is equivalent to moving the right endpoint of original function in opposite direction, thus moving $a$ left will cause $\Delta(S^*)$ to increase and vice versa. □

Now, we are ready to prove Theorem1.

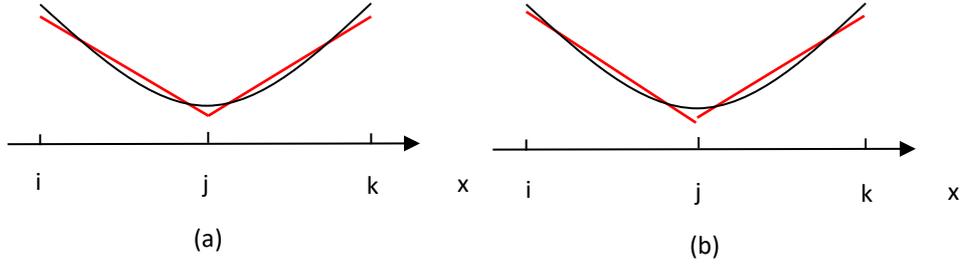

Fig.1 Necessary condition for optimal approximation: the two line segments should have equal optimal approximation errors.

***Proof of Theorem 1***. ***Necessary conditions***. The core idea is to show that if (3.1) does not hold, we can always reduce the approximation errors further. We first consider the case of two segments. As shown in fig.1(a), the segments generated by relu networks are continuous at the intersection point $j$. For arbitrary piecewise linear function $f_2(x)$ that is composed of two segments $S_{ij}$ and $S_{jk}$, the approximation error $\Delta(f_2)$ is

$$\Delta(f_2) = \max(\Delta(S_{ij}), \Delta(S_{jk}))$$

We now cut $f_2(x)$ at point $j$ and adjust the two segments independently. When the three points $i, j$ and $k$ are fixed, apparently we can adjust each segment and make them being the optimal approximations within intervals $[i, j]$ and $[j, k]$ respectively. Let $S^*$ be the segment after adjustment, we have

$$\Delta(S_{ij}) \geq \Delta(S_{ij}^*), \Delta(S_{jk}) \geq \Delta(S_{jk}^*).$$

Therefore,

$$\Delta(f_2) \geq \Delta(f_2') \stackrel{\Delta}{=} \max\left(\Delta(S_{ij}^*), \Delta(S_{jk}^*)\right) \tag{3.13}$$

where $f_2'$ is composed of the two segments $S_{ij}^*$ and $S_{jk}^*$. After adjustment, consider the case where $S_{ij}^*$ and $S_{jk}^*$ are not continuous at point $j$, thus $\Delta(S_{ij}^*) \neq \Delta(S_{jk}^*)$. With loss of generality, we assume $\Delta(S_{ij}^*) > \Delta(S_{jk}^*)$, as shown in fig. 1(b). We then move point $j$ left gradually. With this movement, according to lemma 2, the optimal approximation error $\Delta(S_{ij}^*)$ will decrease monotonically and $\Delta(S_{jk}^*)$ increase monotonically, and hence $\Delta(f_2')$ will decrease by (3.13) and the assumption $\Delta(S_{ij}^*) > \Delta(S_{jk}^*)$. Finally, $\Delta(S_{ij}^*)$ and $\Delta(S_{jk}^*)$ will be equal when $j$ reaches a certain point, say $j_1$. Segments $S_{ij}^*$ and $S_{jk}^*$ will be continuous at $j_1$ due to the fact that endpoints assume optimal approximation errors and $\Delta(S_{ij_1}^*) = \Delta(S_{j_1k}^*)$, thus they are feasible outputs of Relu neural networks. We then stop at $j_1$, and let $f_2^*(x)$ be the final piecewise linear function, we will have

$$\Delta(f_2^*) = \max\Delta(S_{ij_1}^*), \Delta(S_{j_1k}^*)) = \Delta(S_{ij_1}^*) = \Delta(S_{j_1k}^*) \tag{3.14}$$

From (3.13), we also have $\Delta(f_2^*) < \Delta(f_2') \leq \Delta(f_2)$. This indicates $f_2^*(x)$ will have less approximation error than $f_2(x)$. Actually $\Delta(f_2^*)$ is already the optimal approximation error since further movement of $j_1$ will cause its increase. Therefore (3.14) holds when optimal approximation occurs. For the case where $S_{ij}^*$ and $S_{jk}^*$ still intersect at point $j$ after cutting and adjustment, we have $\Delta(S_{ij}^*) = \Delta(S_{jk}^*)$, hence (3.14) already holds. Putting together, we conclude that the approximation errors of the two segments in $f_2^*(x)$ must be equal.

If there are more segments in $f_n(x)$, i.e., $n>2$, we can find the segments $S_I$ with the largest approximation error,

$$\Delta(f_n) = \Delta(S_I) = \max(\Delta(S_1), \Delta(S_2) \ldots \Delta(S_n)),$$

where $I$ is the set containing the indices of segments with the largest approximation error. If there is only one element $i$ in set $I$, we can cut and adjust the left or right endpoint of $S_i$ to make $\Delta(S_i^*) = \Delta(S_{i-1}^*)$ or $\Delta(S_i^*) = \Delta(S_{i+1}^*)$ respectively, just as we did in the $f_2(x)$ case. As such, $\Delta(S_I)$ and consequently $\Delta(f_n)$ are decreased. If there are multiple elements in set $I$, apply the above process sequentially for each element, and consequently $\Delta(f_n)$ is still decreased. We then start the next round of finding $S_I$ and adjustment. This procedure will only stop when all segments have the same approximation error that cannot be decreased anymore. This proves the necessary conditions part of Theorem 1.

***Sufficient conditions***. From the above argument, one can see that once (3.1) holds, any movements of interior endpoints will cause $\Delta(f_n')$ to increase, and subsequent sewing neighboring segments at common endpoints to make $f_n$ be continuous will increase $\Delta(f_n)$ further. Therefore, (3.1) is sufficient for optimal approximation.

***Uniqueness***. We will prove it by contradiction. Assume there are two different optimal piecewise linear functions with the same approximation errors. An illustrative example showing two such functions where $n=3$ is given in Fig. 2.

Geometrically, there must exist two intervals with one of them being included in another, such as $[a, e]$ and $[a, e_1]$ in Fig.2, due to the fact that endpoints $a$ and $b$ are fixed. By the assumption that the two configurations have equal optimal approximation errors and the optimality conditions, one has

$$\Delta(f_3^*) = \Delta(S_{ae}^*), \Delta(f_3^*) = \Delta(S_{ae_1}^*). \tag{3.15}$$

However, $[a, e]$ and $[a, e_1]$ have different lengths, by lemma 2 which claims the monotony of optimal approximation error with respect to interval length, we have

$$\Delta(S_{ae}^*) \neq \Delta(S_{ae_1}^*),$$

which contradicts (3.15). As a result, the optimal approximation $f_n^*(x)$ must be unique. □

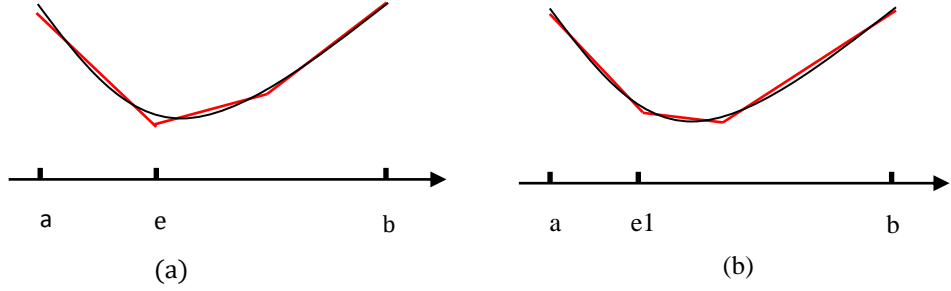

Fig.2 Uniqueness of optimal approximation

### 3.2 Approximation error bounds and approximation rate

We now give our upper and lower bounds of optimal approximation error, along with the approximation rate. The approximation rate measures how fast the optimal approximation error converges with the number of linear segments $n$. We have the following theorem.

**Theorem 2.** *The optimal approximate error $\Delta(f_n^*)$ is bounded as*

$$\frac{(b-a)^2 \cdot \min_{x \in [a,b]} f''(x)}{16} \cdot \frac{1}{n^2} \leq \Delta(f_n^*) \leq \frac{(b-a)^2 \cdot \max_{x \in [a,b]} f''(x)}{16} \cdot \frac{1}{n^2}. \tag{3.16}$$

Theorem 2 implies that $\Delta(f_n^*)$ converges at a rate of $\Theta(1/n^2)$.

***Proof of theorem 2.*** We first prove the upper bound of $\Delta(f_n^*)$ and the lower bound will be obtained in a similar way. For any convex function $f(x)$, we construct a function $\bar{f}(x)$ which has a constant second-order derivative $\bar{f}''(x) = \max_{x \in [a,b]} f''(x)$, i.e., $\bar{f}(x)$ is constructed using the most convex point of $f(x)$. Since $\bar{f}(x)$ is more convex than $f(x)$, when using piecewise linear functions with $n$ segments to approximate them, $\Delta(f_n^*)$ will be upper bounded by the optimal approximation error $\Delta(\bar{f}_n^*)$ of function $\bar{f}(x)$. For $\bar{f}(x)$, (3.2) implies

$$\Delta(f_n^*) = \Delta(S_i^*) \leq \frac{c_i - a_i}{2} \cdot [\bar{f}'(c_i) - \bar{f}'(d_i)], i = 1, 2, \cdots, n, \tag{3.17}$$

where $c_i$ and $d_i$ are determined by function $\bar{f}(x)$. Since $\bar{f}''(x)$ is constant, $c_i$ will be located at the midpoint of $[a_i, b_i]$ by lemma 3 described later in this section, i.e., $c_i - a_i = \frac{b_i - a_i}{2}$, and $d_i$ will be located at midpoint of $[a_i, c_i]$, $d_i - a_i = \frac{c_i - a_i}{2} = \frac{b_i - a_i}{4}$. We have

$$\Delta(f_n^*) \leq \frac{c_i - a_i}{2} \bar{f}'' \cdot [c_i - d_i] = \frac{b_i - a_i}{4} \max_{x \in [a,b]} f''(x) \left[\frac{b_i - a_i}{2} - \frac{b_i - a_i}{4}\right] = \frac{(b_i - a_i)^2}{16} \cdot \max_{x \in [a,b]} f''(x),$$

$$i = 1, 2, \cdots, n. \tag{3.18}$$

Consider the segment corresponding to the shortest interval for which we have $b_i - a_i \leq \frac{b-a}{n}$, thereby we get the following upper bound,

$$\Delta(f_n^*) \leq \frac{(b-a)^2}{16} \cdot \max_{x \in [a,b]} f''(x) \cdot \frac{1}{n^2}.$$

On the other hand, $\Delta(f_n^*)$ is lower bounded by the optimal approximation error of function $\hat{f}(x)$ which has a constant second-order derivative $\hat{f}'' = \min_{x \in [a,b]} f''(x)$, thus

$$\Delta(f_n^*) = \Delta(s_i^*) \geq \frac{c_i - a_i}{2} \cdot [\hat{f}'(c_i) - \hat{f}'(d_i)], i = 1, 2, \cdots, n. \tag{3.19}$$

Consider the segment corresponding to the largest interval for which $b_i - a_i \geq \frac{b-a}{n}$ holds, we get

$$\Delta(f_n^*) \geq \frac{b_i - a_i}{4} \cdot \min_{x \in [a,b]} f''(x) \left[ \frac{b_i - a_i}{2} - \frac{b_i - a_i}{4} \right] = \frac{(b_i - a_i)^2}{16} \cdot \min_{x \in [a,b]} f''(x) \geq$$

$$\frac{(b-a)^2}{16} \min_{x \in [a,b]} f''(x) \cdot \frac{1}{n^2}. \tag{3.20}$$

This completes the proof of lower bound. □

For functions $f(x)$ with constant second-order derivatives, $\max_{x \in [a,b]} f''(x) = \min_{x \in [a,b]} f''(x)$, hence there will be no gap between the upper and lower bounds, and both of them equal the exact optimal approximation error. In section 5.3, we will present experimental results for function $f(x) = x^2$ which has a constant second-order derivative, and compare the optimal approximation error with our theoretical bounds.

**Lemma 3**. *When using line segments to approximate convex functions $f(x)$ that have constant second-order derivatives, c will be located at the midpoint of [a, b] and d located at the midpoint of [a, c].*

***Proof.*** Since $f''(x)$ is constant, we have $f = \frac{f''}{2}x^2 + kx + e$, where $k$ and $e$ are real-valued constants. This implies $f' = f'' \cdot x + k$. By definition in (3.3), $f'(c) = \frac{f(b) - f(a)}{b - a} = \frac{\frac{f''}{2}b^2 + kb - \left(\frac{f''}{2}a^2 + ka\right)}{b - a} = f'' \cdot \frac{(b+a)}{2} + k = f'' \cdot c + k$, therefore $c = \frac{b+a}{2}$.

The fact that $d = \frac{c+a}{2}$ can be proved similarly. □

**Remark**: Besides giving the exact optimal approximation error in (3.2), we also give its lower and upper bounds in (3.16). [8, 9, 10] have given lower bounds of approximation errors as well. All these lower bounds, including ours, are derived using the local or global convexity of target functions, and are all proportional to $\frac{1}{n^2}$ (this can be obtained through careful examination of the derivations in these works), indicating these bounds are at the same level. For example, the lower bound in theorem 6 of [9] is $\frac{(b_i - a_i)^2}{16} \cdot \min_{x \in [a,b]} f''(x)$ (where we have rewritten it using notations of this paper), which is exactly our lower bound in (3.20). The bounds in [8] and [10] are not directly comparable with ours. For instance, the approximation error in [10] is defined with $L_2$ norm.

## 4 Relu neural network architectures that achieve optimal approximations
### 4.1 The Relu neural network architecture

In this section, given the optimal approximation $f_n^*(x)$, we will present Relu network architectures that can generate the segments required in $f_n^*(x)$. We will generate the segments one by one.

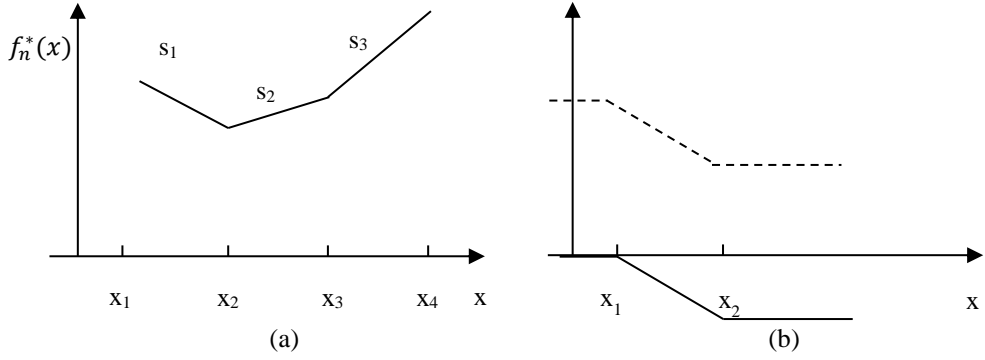

Fig.3 Generating linear segments by Relu neural networks

The process is illustrated in Fig.3. Each segment can be generated by a two-hidden-layer Relu network. The output of this two-hidden-layer network for segment $S_i$ is as follows,

$$O_i = sgn(k_i) \cdot \sigma(|k_i| \cdot [\sigma(x - x_i) - \sigma(x - x_{i+1})]), \tag{4.1}$$

where $k_i$ is the slope of segment $S_i$, and $sgn(k_i)$ is the sign of $k_i$. $O_i$ can be expanded as follows according to (4.1),

$$O_i = \begin{cases} 0, & x < x_i \\ k_i(x - x_i), & x_i \leq x < x_{i+1} \\ k_i(x_{i+1} - x_i), & x \geq x_{i+1}. \end{cases} \tag{4.2}$$

The real line in fig.3(b) shows $O_1$ for segment $S_1$. If we move $O_1$ upwards by $f_n^*(x_1)$, we can obtain $O_1' = f_n^*(x_1) + O_1$ which perfectly coincides with $S_1$ in $[x_1, x_2]$, as shown by the dotted line in Fig.3(b). The subsequent segments are generated in similar ways. Finally, we sum them up and get

$$\widetilde{f}_n(x) = f_n^*(x_1) + \sum_{i=1}^n O_i. \tag{4.3}$$

$\widetilde{f}_n(x)$ equals $f_n^*(x)$ everywhere in $[x_1, x_{n+1}]$, which is the interval of interest.

$\widetilde{f}_n(x)$ can be implemented as a network architecture that consists a input, $n$ parallel modules of sub-network that computes $O_i$ and a output neuron performing the final summation in (4.3). Fig.4(a) shows this architecture. Ignoring the input and output neurons, there are totally $3n$ neurons in it.

The above architecture is a fixed-depth one. However, we can transform it into a fixed-width architecture with variable depth. In order to do so, we define a width-5 layer with elements ($x$, $\sigma(x - x_i), \sigma(x - x_{i+1}), O_i, f_n^*(x)$) and use four such layers to generate one segment. Fig.4(b) shows the outputs of each layer and connections between layers for producing segment $S_1$. Ignore possible reuse, in total 20 neurons are required to generate each segment. Therefore, there are $20n$ neurons in this fixed-width architecture.

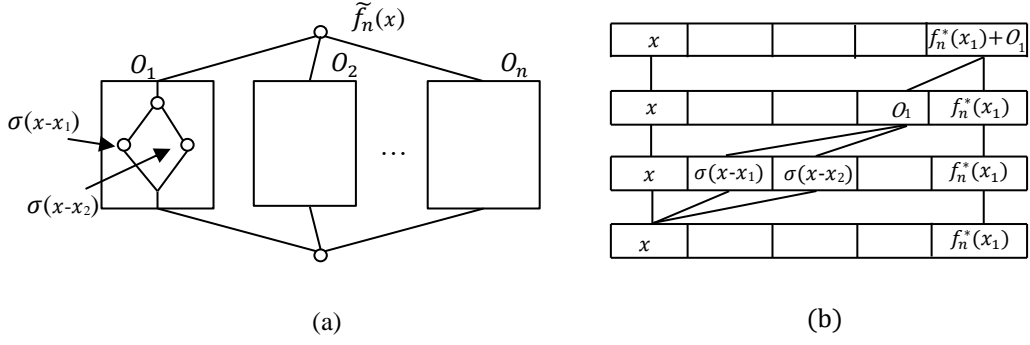

Fig.4 Relu neural network architectures that produce optimal approximations

**Remark**: Fixed-width architectures have been studied in [3, 4], and a fixed-depth architectures is given in [5]. However, these architectures did not consider optimal approximations. Another related fixed-depth Relu network architecture appeared in [17], which also uses addition of linear pieces to get the final output. However, [17] adds multiple 2-pieces piecewise linear functions, and hence has only one hidden layer. Furthermore, [17] do prove the existence of such Relu network architecture, but do not give an explicit expression of the component functions. On the contrary, we give in (4.1) an explicit construction of each component, which is a two-hidden-layer sub-network.

**4.2 Approximation error bounds with respect to network size**

We now give the upper and lower bounds of optimal approximation error with respect to network size, as expressed by the following corollary.

**Corollary 1.** *Given a Relu network of N neurons and L layers (including the input and output layers), and any convex function $f(x)$, the optimal approximation error is upper bounded by $\frac{9(b-a)^2}{16} \cdot \max_{x \in [a,b]} f''(x) \cdot \frac{1}{N^2}$ and the lower bounded by $\frac{(b-a)^2}{16 \cdot 2^{2L-4}} \cdot \min_{x \in [a,b]} f''(x) \cdot \frac{1}{N^{2L-4}}$ .*

*Proof.* Using the fixed-depth architecture shown in Fig.4(a), one has $N = 3n$. By substituting $n = \frac{N}{3}$ into the upper bound in (3.16), we get the desired upper bound.

By lemma 4 of (Yarotsky, 2017), the number of linear pieces generated by a relu network is upper bounded as follows,
$$n \leq (2N)^{L-2}.$$
Combining with the lower bound in (3.16), the lower bound of approximation error in terms of $N$ and $L$ is obtained. □

# 5 An algorithm to find optimal approximations and experimental results
## 5.1 Finding optimal approximations

Theorem 1 gives the necessary and sufficient conditions for optimal approximations of convex functions $f(x)$ with piecewise linear functions $f_n(x)$. In the proof of Theorem 1, we choose the segment with the largest approximation error and, by moving its endpoints appropriately we can always reduce approximation error $\Delta(f_n)$, unless all segments have equal approximation errors.

However, the segments with low approximation errors remain intact during this process and hence infinite number of iterations is needed to converge. In this section, in order to promote convergence, we design an algorithm which adjusts all segments during each iteration.

Given convex functions $f(x)$, and the number of line segments $n$, we design the following algorithm 1 to find optimal approximations.

---

Input: number of segments $n$; target function $f(x)$; stepsize
Output: optimal linear segments; $\{\Delta(S_i)\}$

---

Initialization of intervals;

// notation: $\Delta(S_{max}) = \max\limits_{i \in \{1,2,\cdots,n\}} \Delta(S_i)$, $\Delta(S_{min}) = \min\limits_{i \in \{1,2,\cdots,n\}} \Delta(S_i)$

do{    // a round
    for ($i = 1$ to $n$-1) {    // adjust interior endpoints in sequence
        compute $\Delta(S_i^*)$ and $\Delta(S_{i+1}^*)$;
        if ($\Delta(S_i^*) > \Delta(S_{i+1}^*)$){
            while ($\Delta(S_i^*) > \Delta(S_{i+1}^*)$){
                move the common endpoint of $S_i$ and $S_{i+1}$ left by stepsize;
                recompute $\Delta(S_i^*)$ and $\Delta(S_{i+1}^*)$ ;
            }
        }elseif ($\Delta(S_i^*) < \Delta(S_{i+1}^*)$){
            while ($\Delta(S_i^*) < \Delta(S_{i+1}^*)$){
                move the common endpoint of $S_i$ and $S_{i+1}$ right by stepsize;
                recompute $\Delta(S_i^*)$ and $\Delta(S_{i+1}^*)$ ;
            }
        }
    }
}while( ($\Delta(S_{max})$-$\Delta(S_{min})$ in previous round)  >  ($\Delta(S_{max})$-$\Delta(S_{min})$ in current round) );
Output the optimal linear segments and $\{\Delta(S_i)\}$ ;

Algorithm 1. An algorithm to find optimal approximations

The initialization of intervals can be arbitrary, and the most simple one is to evenly distribute the intervals, i.e., $a_i = a + \frac{b-a}{n} \cdot (i-1)$, $b_i = a + \frac{b-a}{n} \cdot i$ ($i = 1, 2, \cdots, n$). Each segment is then optimized independently. During each round, we adjust the interior endpoints consecutively by repeatedly moving each of them with a small stepsize, and the purpose of each adjustment is to make two neighboring segments have almost equal approximation errors.

### 5.2 Proof of convergence

We now proceed to prove that the Algorithm 1 converges, i.e., the gap between the largest and smallest approximation errors diminishes after each round. More formally, we have the following theorem.

**Theorem 3.** $\Delta(S_{max}) - \Delta(S_{min})$ *diminishes after each round.*

*Proof.* We prove by induction. During each round, first consider adjusting the first two segments $S_1$ and $S_2$, and suppose $\Delta(S_1) \neq \Delta(S_2)$. Now we move their common endpoint gradually to reduce $\max(\Delta(S_1), \Delta(S_2))$. At the end of the movement, denote by $S_i'$ the $i$th segment after adjustment, we will have $\Delta(S_1') = \Delta(S_2')$ and hence $\min(\Delta(S_1), \Delta(S_2)) < \Delta(S_1') < \max(\Delta(S_1), \Delta(S_2))$. We denote by $\Delta\left(S_{max \atop 1 \sim i}\right)$ the maximal approximation error among all segments in $\{S_1, S_2, \cdots, S_i\}$ considered so far, and by $\Delta\left(S'_{max \atop 1 \sim i}\right)$ the maximal approximation error among segments in $\{S_1', S_2', \cdots, S_i'\}$. For $S_1$ and $S_2$, the following holds after adjustment,

$$\Delta\left(S_{max \atop 1 \sim 2}\right) - \Delta\left(S'_{max \atop 1 \sim 2}\right) = \max(\Delta(S_1), \Delta(S_2)) - \Delta(S_1') > 0. \tag{5.1}$$

This case is shown in Fig.5(a).

Suppose for the $i$th segment, we already have $\Delta\left(S_{max \atop 1 \sim (i-1)}\right) - \Delta\left(S'_{max \atop 1 \sim (i-1)}\right) > 0$. We then want to prove $\Delta\left(S_{max \atop 1 \sim i}\right) - \Delta\left(S'_{max \atop 1 \sim i}\right) > 0$. There are two possible cases.

*Case a*. The ($i$-1)th segment $S'_{i-1}$ is the one with the maximal optimal approximation error after previous adjustments, i.e., $\Delta(S'_{i-1}) = \Delta\left(S'_{max \atop 1 \sim (i-1)}\right)$. We have $\Delta\left(S_{max \atop 1 \sim i}\right) = max(\Delta(S_i), \Delta\left(S_{max \atop 1 \sim (i-1)}\right))$, and after adjusting the common endpoint of $S_i$ and $S'_{i-1}$, we also have

$\Delta\left(S'_{max \atop 1 \sim i}\right) = max(max(\Delta(S_i), \Delta(S'_{i-1})) - \delta, \max_{j=1,2,\cdots,i-2} \Delta(S'_j))$, where $\delta$ is the reduction of optimal approximation error caused by moving the common endpoint of $S_i$ and $S'_{i-1}$. Therefore,

$$\Delta\left(S_{max \atop 1 \sim i}\right) - \Delta\left(S'_{max \atop 1 \sim i}\right) = max\left(\Delta(S_i), \Delta\left(S_{max \atop 1 \sim (i-1)}\right)\right) - max\left(max(\Delta(S_i), \Delta(S'_{i-1})) - \delta, \max_{j=1,2,\cdots,i-2} \Delta(S'_j)\right) = min(max(\Delta(S_i), \Delta\left(S_{max \atop 1 \sim (i-1)}\right)) - max(\Delta(S_i), \Delta(S'_{i-1})) + \delta, max(\Delta(S_i), \Delta\left(S_{max \atop 1 \sim (i-1)}\right)) - \max_{j=1,2,\cdots,i-2} \Delta(S'_j)). \tag{5.2}$$

Using the inductive hypothesis $\Delta\left(S_{max \atop 1 \sim (i-1)}\right) > \Delta\left(S'_{max \atop 1 \sim (i-1)}\right) = \Delta(S'_{i-1})$ and the fact $\delta > 0$, we have

$$max\left(\Delta(S_i), \Delta\left(S_{max \atop 1 \sim (i-1)}\right)\right) - max(\Delta(S_i), \Delta(S'_{i-1})) + \delta \geq \delta > 0. \tag{5.3}$$

Note that $\Delta\left(S_{max \atop 1 \sim (i-1)}\right) > \Delta\left(S'_{max \atop 1 \sim (i-1)}\right) > \max_{j=1,2,\cdots,i-2} \Delta(S'_j)$, thus

$$max\left(\Delta(S_i), \Delta\left(S_{max \atop 1 \sim (i-1)}\right)\right) - \max_{j=1,2,\cdots,i-2} \Delta(S'_j) > 0. \tag{5.4}$$

Combination of (5.2), (5.3) and (5.4) yields

$$\Delta\left(S_{max \atop 1 \sim i}\right) - \Delta\left(S'_{max \atop 1 \sim i}\right) > 0. \tag{5.5}$$

Fig.5(b) shows case *a* before moving the common endpoint of $S_i$ and $S'_{i-1}$.

*Case b.* the ($i$-1)th segment $S'_{i-1}$ is not the one with the maximal optimal approximation error. Therefore, the segment with maximal optimal approximation error must be included in $\{S'_1, S'_2, \cdots, S'_{i-2}\}$, and as a result we have $\Delta\left(S'_{\substack{max\\1\sim(i-1)}}\right) = \max_{j=1,2,\cdots,i-2} \Delta(S'_j) > \Delta(S'_{i-1})$. Note that for case b, (5.2) still holds, and we also have $\Delta\left(S_{\substack{max\\1\sim(i-1)}}\right) > \Delta\left(S'_{\substack{max\\1\sim(i-1)}}\right) > \Delta(S'_{i-1})$ and $\Delta\left(S_{\substack{max\\1\sim(i-1)}}\right) > \Delta\left(S'_{\substack{max\\1\sim(i-1)}}\right) = \max_{j=1,2,\cdots,i-2} \Delta(S'_j)$, hence we can conclude that (5.3) and (5.4) still hold as well for case *b*, which again leads to (5.5). Fig.5(c) illustrates case b before the adjustment.

Above we have proved that the current maximal optimal approximation error $\Delta\left(S_{\substack{max\\1\sim i}}\right)$ will diminish after adjustment. Similarly, we can prove that the current minimal optimal approximation error $\Delta\left(S_{\substack{min\\1\sim i}}\right)$ will increase after adjustment. We omit the detail to save space. Combining (5.1) and (5.5), we conclude by induction that $\Delta(S_{max}) - \Delta(S_{min})$ diminishes after each round.

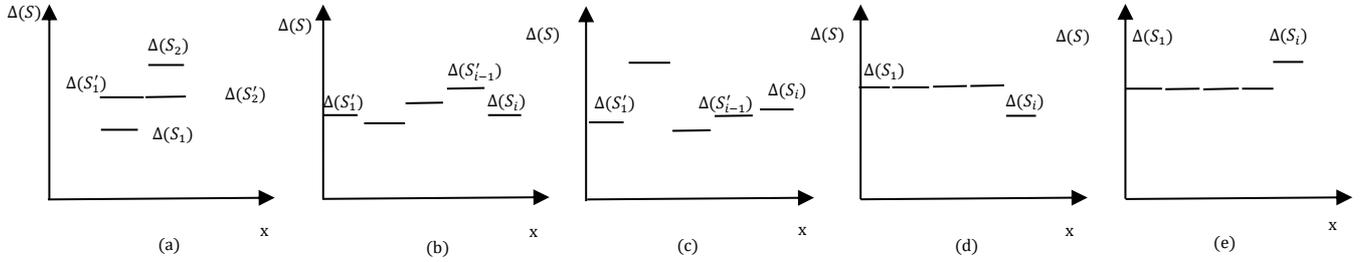

Fig. 5 The convergence of Algorithm 1 under different circumstances

It is worthy to point out that if all previous segments $\{S_1, S_2, \cdots, S_{i-1}\}$ have the same approximation errors, and if $\Delta(S_i)$ is smaller than them, as shown in Fig.5(d), then $max(\Delta(S_i), \Delta\left(S_{\substack{max\\1\sim(i-1)}}\right)) - \max_{j=1,2,\cdots,i-2} \Delta(S'_j) = 0$ due to the fact that no adjustments happen for all previous segments in $\{S_1, S_2, \cdots, S_{i-1}\}$. According to (5.2), $\Delta\left(S_{\substack{max\\1\sim i}}\right) - \Delta\left(S'_{\substack{max\\1\sim i}}\right)$ will be zero, i.e., no diminish happens. Similarly, for the case shown in Fig.5(e), there will be no increase of minimal optimal approximation error. However, increase of minimal error still happens for the case of Fig.5(d), and decrease of maximal error still happens for the case of Fig.5(e). One of these two cases must occur, otherwise all segments in $\{S_1, S_2, \cdots, S_n\}$ will have the same approximation errors, and there will be no adjustment anymore and convergence has been reached already. Therefore, the gap between maximal and minimal optimal approximation errors still gets smaller until convergence. □

### 5.3 Experimental Results

In order to demonstrate the effect of algorithm 1 and see how the optimal approximations look like, we implement Algorithm 1 in Python and then experiment with three different functions: $e^x, x^2, x^3$ using a commodity laptop computer. The interval of interest is $[0,1]$ for $e^x$ and $x^3$, and $[-1,1]$ for $x^2$. These functions are strict convex in the specified intervals. The number of segments *n* is set to 2, 3, 5 and 10 respectively. During initialization, the sub-intervals are evenly distributed, and each segment is optimized independently using (3.6). The stepsize is set to $10^{-5}$.

Table 1 gives our experimental results for these functions with different number of line segments. The mean approximation errors, theoretical lower and upper bounds of approximation errors computed by Theorem 2, gaps between maximal and minimal approximation errors, number of rounds to converge and running times in seconds are presented in this table. One can see that the gaps are very small compared with the mean errors, indicating that the sufficient and necessary conditions in Theorem 1 are reached and convergence occurs. Fig.6 shows the final approximation effects, exhibiting that not only the optimal line segments have equal approximation errors, but also they are connected at shared endpoints and thus realizable with Relu neural networks.

Table 1 clearly shows that the mean approximation errors are within the theoretical lower and upper bounds, thus demonstrating their effectiveness. Function $f(x) = x^2$ has constant second-order derivative, hence its lower and upper bounds are equal. One can see from table 1 that for this function, the mean optimal approximation errors obtained by Algorithm 1 are exactly the same with theoretical bounds.

Table 1. Experimental results of optimal approximations with different number of linear pieces

| Target function | $n$ | $\frac{1}{n}\sum_{i=1}^{n}\Delta(S_i)$ | theoretical upper bound | theoretical lower bound | $\max_i\Delta(S_i) - \min_i\Delta(S_i)$ | #round | running time (s) |
|---|---|---|---|---|---|---|---|
| $e^x$ | 2 | 0.02635 | 0.04247 | 0.01563 | $1.1356*10^{-6}$ | 6177 | 0.6647 |
| | 3 | 0.01170 | 0.01886 | 0.00694 | $1.4222*10^{-6}$ | 11028 | 1.2448 |
| | 5 | 0.00421 | 0.00680 | 0.00250 | $6.5323*10^{-7}$ | 19896 | 2.1884 |
| | 10 | 0.00105 | 0.00171 | 0.00063 | $3.5872*10^{-7}$ | 41065 | 4.6130 |
| $x^2$ | 2 | 0.125 | 0.125 | 0.125 | 0 | 1 | 0.00026 |
| | 3 | 0.05556 | 0.05556 | 0.05556 | 0 | 1 | 0.00069 |
| | 5 | 0.02000 | 0.02000 | 0.02000 | $2.0*10^{-9}$ | 9 | 0.00252 |
| | 10 | 0.00500 | 0.00500 | 0.00500 | $2.776*10^{-17}$ | 18 | 0.00080 |
| $x^3$ | 2 | 0.04486 | 0.09375 | 0 | $2.6875*10^{-6}$ | 11546 | 1.2748 |
| | 3 | 0.01946 | 0.04167 | 0 | $3.1486*10^{-6}$ | 22293 | 2.4536 |
| | 5 | 0.00687 | 0.01500 | 0 | $7.6881*10^{-7}$ | 43065 | 4.6004 |
| | 10 | 0.00169 | 0.00375 | 0 | $1.1087*10^{-6}$ | 93872 | 10.1207 |

## 6 Conclusion

We have considered the optimal approximation problem in this paper: given a convex function $f(x)$ and the number of approximating linear pieces *n*, what is the minimal $L_\infty$ approximation error? We give the necessary and sufficient conditions for optimality and prove its uniqueness. Lower and upper bounds of optimal approximation error are given, showing that the approximation error converges at a rate of $\Theta(1/n^2)$. We then design a constant-depth and a constant-width Relu network architectures to generate these optimal linear pieces. Finally, we propose an algorithm to search for optimal approximations and prove its convergence, and validate its effectiveness with experimental results.

There are some promising directions for future work. We want to explore the optimal approximation of more general functions beyond convex ones. Our current optimal approximation relys on one-dimensional movement of common endpoint of two adjacent segments, thus only applies to univariate functions. We plan to extend the optimal approximation to functions with higher-dimensional inputs. Lastly, each optimal segment is currently independently output by Relu neural networks, how to improve the optimal approximation error by exploiting the internal dependence within each network deserves in-depth exploration.

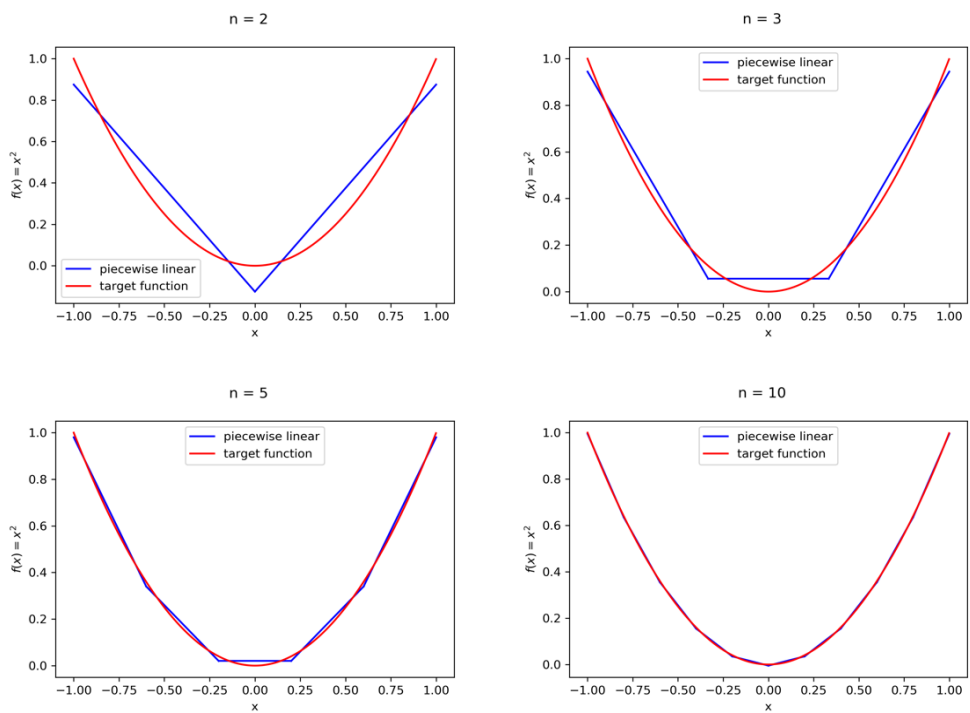

(a) $f(x) = x^2$

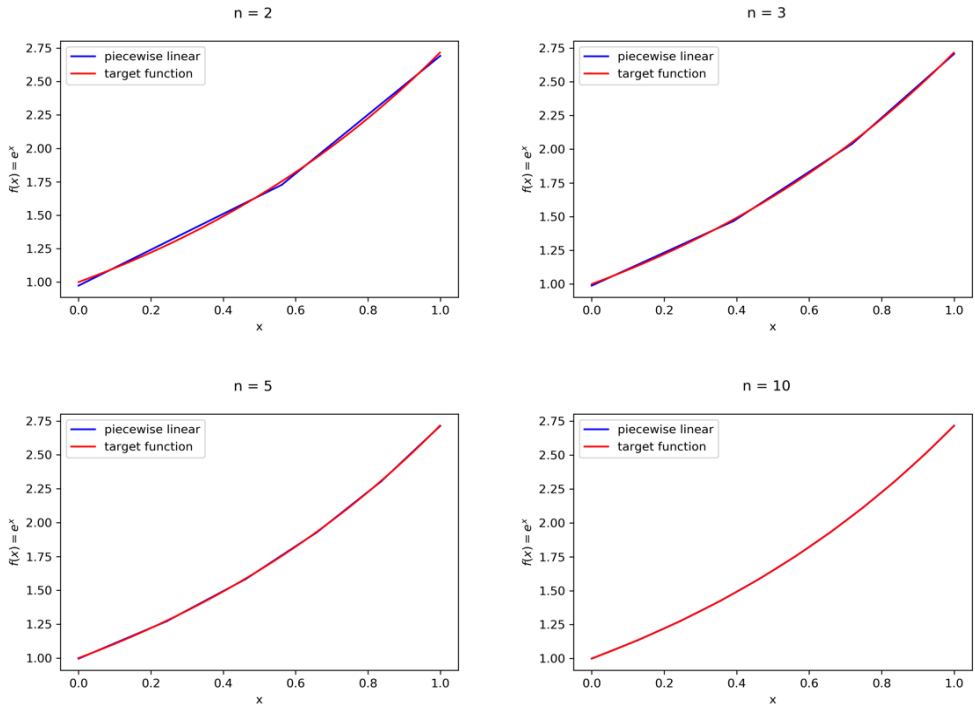

(b) $f(x) = e^x$

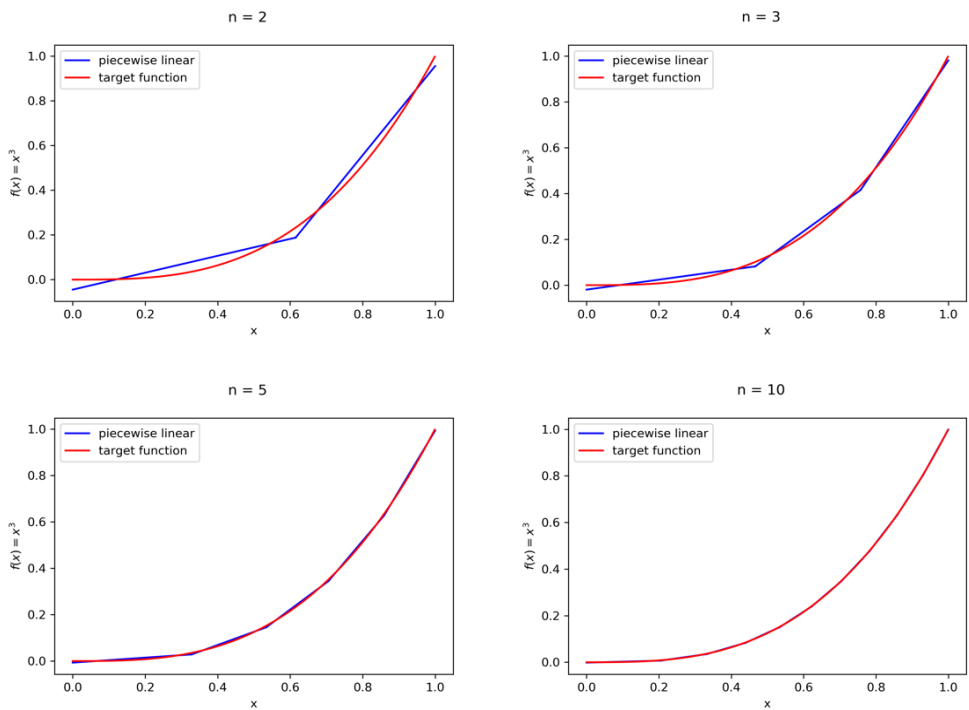

(c) $f(x) = x^3$

Fig. 6 Optimal approximations obtained with Algorithm 1